\documentclass[runningheads]{llncs}
\usepackage[T1]{fontenc}
\usepackage{amsmath}
\usepackage{xspace}
\usepackage{booktabs}
\usepackage{xcolor}
\usepackage{pifont}
\newcommand{\cmark}{\ding{51}}  

\usepackage{algpseudocode}

\usepackage{graphicx}

\usepackage[breaklinks,colorlinks]{hyperref}
\usepackage{color}

\newcommand{\method}{Fake3DGS\xspace}

\begin{document}

\title{\method: A Benchmark for 3D Manipulation Detection in Neural Rendering}
%
%
\author{Davide Di Nucci \and
Riccardo Catalini \and 
Guido Borghi \and
Roberto Vezzani}
\authorrunning{D. Di Nucci et al.}
%
\institute{University of Modena and Reggio Emilia, Via Pietro Vivarelli, 41121 Modena, Italy
\email{name.surname@unimore.it}\\}
\titlerunning{Fake3DGS}

\maketitle              
\begin{abstract}
Recent advances in 3D reconstruction and neural rendering, particularly 3D Gaussian Splatting, make it feasible and simple to edit 3D scenes and re-render them as highly realistic images. Therefore, security concerns arise regarding the authenticity of 3D content. 
Despite this threat, 3D fake detection remains largely unexplored in the literature, and most existing work is limited to 2D space.
Therefore, in this paper, we formalize the concept of 3D fake detection and introduce Fake3DGS, a dataset of 3D Gaussian splatting scenes and corresponding rendered views, where fake images are produced by controlled manipulations of geometry, appearance, and spatial layout, while preserving high visual realism. Using this benchmark, we demonstrate that current state-of-the-art 2D detectors struggle to distinguish between original and 3D manipulated images. To bridge this gap, we introduce a 3D-aware detection method that leverages multi-view coherence and features derived from the Gaussian splatting representation. Experimental results demonstrate a substantial improvement in recognizing modified 3D content, underscoring the validity of the new dataset and the necessity for authenticity assessment techniques that extend beyond 2D evidence. Code and data are publicly released\footnote{\url{https://github.com/iot-unimore/Fake3DGS}} for future investigations.
\end{abstract}

\section{Introduction}

With the widespread adoption of Generative AI, 2D deep fake detection has become a well-established task in computer vision, with numerous datasets and benchmarks available~\cite{Celeb_DF_cvpr20,wang2023dire,baraldi2024contrastive,dang2020detection}. However, generative methods are rapidly evolving beyond the two-dimensional domain. Recent developments integrate models such as Variational Autoencoders (VAE)~\cite{kingma2013auto}, Generative Adversarial Networks (GANs)~\cite{goodfellow2014generative}, and Diffusion Models~\cite{rombach2022high} to improve 3D reconstruction fidelity and reduce inference time. For instance, modern text-to-3D pipelines rely on text-to-image generation followed by 3D synthesis to ensure both semantic consistency and geometric accuracy~\cite{xu2023dream3d,liu2024pi3d}. Other approaches enable complex edits on real 3D scenes~\cite{he2024customize,wu2024gaussctrl}.

Unlike 2D manipulations~\cite{borghi2021automated}, edits performed directly in the 3D scene representation can produce an unlimited number of photorealistic, view-consistent images, making traditional artifact-based detection strategies limited in efficacy. This shift introduces a new class of deep fakes where geometric consistency becomes an advantage for the attacker rather than a vulnerability. The ability to edit 3D scenes is increasingly deployed in Augmented Reality or Virtual Reality environments, digital twins, and simulation pipelines for autonomous systems~\cite{jiang2024vrgs,zhou2024drivinggaussian,zhou2024simgen}.
These elements make the authenticity of 3D content not only a scientific challenge but a security concern. Indeed, despite these progresses and risks, 3D fake detection remains largely unexplored in the literature. Also, while 2D deep fake detection benefits from mature benchmarks and shared evaluation protocols, the 3D domain still lacks definitions, datasets, and standardized procedures for assessing scene authenticity. 

Motivated by these reasons, we formalize 3D fake detection as the task of determining whether a 3D scene representation has been altered in its geometry, spatial layout, or appearance parameters, regardless of how realistic its rendered images appear. We establish the first benchmark for authenticity assessment in 3D Gaussian Splatting~\cite{kerbl20233d}, laying the foundations for future research on multi-view consistency and 3D-aware forensics.

In summary, the novel contributions of this work are the following:
\begin{itemize}
    \item The 3D deep fake detection task is formalized and defined.
    \item A new dataset, called {\method} has been generated, annotated and published to allow fair comparisons and to benchmark current and future techniques; the dataset contains original and modified 3D scenes directly stored  as 3D Gaussian Splatting representations.
    \item A specific detector is proposed and compared with common solutions for fake image detection, showing promising performances and, at the same time, demonstrating the need of specific techniques to solve the task.  
 \end{itemize}

\section{Related work}
\subsection{3D Novel View Synthesis}
Novel view synthesis focuses on producing photorealistic images of a 3D scene from previously unseen camera viewpoints. While early methods based on Neural Radiance Fields (NeRF)~\cite{mildenhall2021nerf} demonstrated that continuous radiance fields with differentiable volume rendering can achieve high-quality results, subsequent works mainly improved efficiency, \textit{e.g.}, via multiresolution hash encodings~\cite{muller2022instant} or explicit voxel grids storing density and spherical harmonics~\cite{fridovich2022plenoxels}.

3D Gaussian Splatting~\cite{kerbl20233d} has emerged as a compelling alternative to NeRF, offering significant gains in both training and rendering speed while preserving high image quality~\cite{di2025brum}. To mitigate its remaining limitations, a rapidly growing body of work explores complementary research directions. One line of work targets compactness and deployment on resource-constrained hardware by compressing the Gaussian representation to reduce storage and memory overhead~\cite{morgenstern2024compact,niedermayr2024compressed,zhang2025gaussianimage}. Another focuses on text-to-3D scene generation has quickly emerged as a prominent application, combining 3D Gaussian Splatting with powerful 2D diffusion priors~\cite{zhou2025dreamscene360,chen2024text,liang2024luciddreamer,li2024controllable,igs2gs}. In parallel, 3D scene decomposition and segmentation to enable fine-grained editing operations~\cite{cen2023segment,hu2024semantic,chen2024gaussianeditor,li2024controllable}.

We further observe that realistically edited 3D Gaussian scenes are scarce, and there is currently no standardized way to evaluate editing methods in this setting. To fill this gap, we introduce a dedicated benchmark for 3D Gaussian scene editing, enabling systematic and reproducible comparison across existing and future approaches.

\subsection{Security and manipulation for 3D Gaussian Splatting}
3D Gaussian Splatting (3DGS)~\cite{kerbl20233d} introduces a new security threat model: an adversary can directly edit Gaussian parameters to produce view-consistent yet deceptive renderings, enabling malicious scene manipulations that may not leave obvious 2D artifacts. Recent work has analyzed this attack surface and demonstrated both training-time and post-training manipulations of 3DGS models~\cite{hull2025vulnerabilities}. 

A complementary defense direction focuses on tamper-evidence and provenance via watermarking of 3DGS assets, embedding signals in the model parameters and/or rendered outputs to enable later verification under distortions and model transformations~\cite{chen2025guardsplat,jang2025_3dgsw,huang2025marksplatter,in2025compmarkgs}. In contrast to watermark-based verification (active protection), our work addresses \emph{passive} 3D manipulation detection under realistic editing pipelines by introducing a large-scale benchmark and a scene-level detector operating on Gaussian-derived features.

\subsection{Fake detection}
Fake image detection has rapidly evolved into a key research topic in recent years. Early efforts \cite{rossler2019faceforensics++,wang2019fakespotter,borghi2021double} concentrated mainly on identifying synthetic human faces and reverse-engineering the fingerprints of specific GAN generators. A complementary line of work explicitly targets perceptible artifacts, such as facial asymmetries, geometric inconsistencies, or implausible shadows and lighting \cite{farid2022lighting,farid2022perspective,matern2019exploiting}. However, as image synthesis models have advanced, these obvious defects are quickly reduced and now rarely appear in state-of-the-art generative pipelines, which limits the effectiveness of methods that rely solely on visible cues. Building on the GAN-based setting, subsequent studies \cite{cozzolino2018forensictransfer,gragnaniello2021gan,wang2020cnn} have examined the zero-shot generalization ability of detectors when confronted with previously unseen generators, showing that different GAN architectures often share exploitable common artifacts. In parallel, Frank et al. \cite{frank2020leveraging} investigated the frequency domain, revealing systematic spectral discrepancies between real and generated images.

The advent of Diffusion Models~\cite{rombach2022high} has shifted the focus of the community. Although diffusion-based generators exhibit their own characteristic generation signatures \cite{corvi2023intriguing}, recent work \cite{corvi2023detection} has demonstrated that detectors trained exclusively on GAN outputs transfer poorly to these newer models. This observation has spurred a wave of approaches designed specifically for diffusion-generated content \cite{ojha2023towards,sha2023fake}, many of which build on CLIP \cite{radford2021clip} as the underlying feature space. A distinct strategy is proposed by Wang et al. \cite{wang2023dire}, who detect forgeries by analyzing the difference between an input image and its reconstruction obtained from a pre-trained diffusion model. 

However, all the above methods operate purely in 2D space, implicitly assuming that fake content can be revealed by pixel- or texture-level artifacts alone. This assumption breaks down in the presence of advanced 3D pipelines such as 3D Gaussian Splatting, which enable realistic edits to geometry, appearance, and layout. As we show in our experiments, state-of-the-art deepfake detectors struggle to distinguish original from 3D-edited views in this setting. To address this gap, we introduce Fake3DGS, a specific dataset of edited 3D Gaussian scenes for benchmarking image authenticity in a 3D manipulation context.

\section{\method Dataset}
\begin{figure}[h]
    \centering
    \includegraphics[page=1, width=1\linewidth]{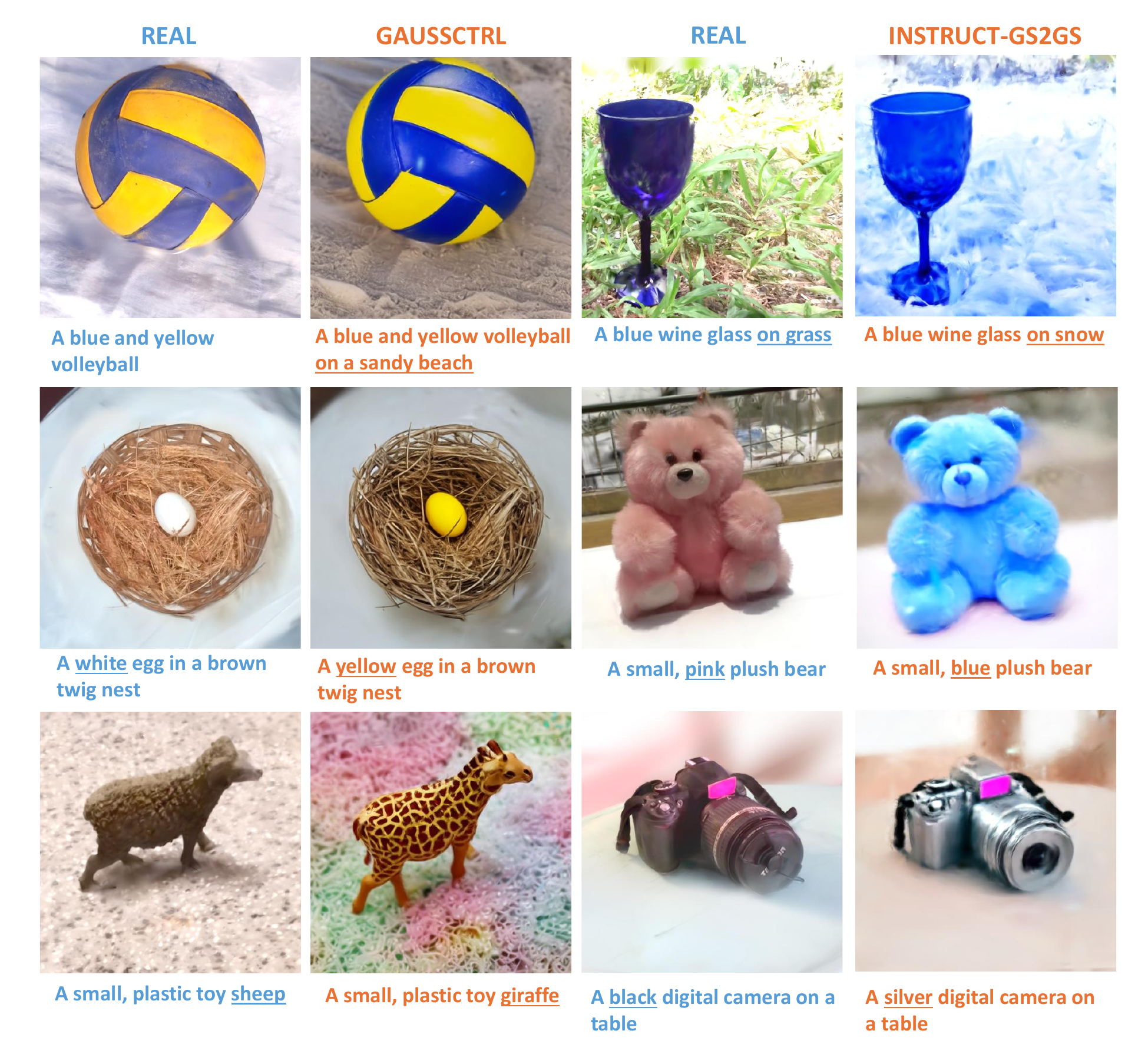}
    \caption{Sample renderings from the {\method} Dataset. Below each view of the scene, the corresponding prompt used to generate edited samples is reported.}
    \label{fig:renderings}
\end{figure}

To evaluate 3D fake detection in a controlled yet realistic setting, we introduce \method dataset, a large-scale benchmark of original and edited 3D Gaussian Splatting scenes. The dataset comprises more than 41k reconstructed scenes, evenly balanced between real and manipulated samples. 

We export all reconstructions as \texttt{nerfstudio}\footnote{\url{https://github.com/nerfstudio-project/nerfstudio}} checkpoints, which provides a widely adopted and well-documented format for neural rendering research, making the dataset easier to use and reproduce with standard tooling. We then post-process each 3DGS model using the Gaussian splatting compression strategy of Self-Organizing Gaussian Grids (SOGS)~\cite{morgenstern2024compact}, leveraging the \texttt{gsplat}\footnote{\url{https://github.com/nerfstudio-project/gsplat}} implementation~\cite{ye2025gsplat} that is integrated in the \texttt{nerfstudio} ecosystem. Concretely, we independently quantize Gaussian parameters with uniform scalar quantization (8-bit per parameter in our implementation). The quantized tensors are then reorganized into locally coherent 2D grids (via sorting) and encoded with lossless PNG compression, enabling exact recovery of the quantized values at load time. This post-processing substantially reduces storage and I/O overhead, shrinking the dataset footprint from approximately $7$\,TB to about $200$\,GB, which makes large-scale training and evaluation feasible on standard research hardware while preserving faithful reconstruction of the compressed representation.

Figure~\ref{fig:renderings} shows representative examples from \method dataset. For each underlying scene we report a real rendering and its edited counterpart generated with either GaussCtrl~\cite{wu2024gaussctrl} or Instruct-GS2GS~\cite{igs2gs} (see Sect.~\ref{sec:generation}). The edits cover the three instruction families used in our benchmark background/surface changes (\textit{e.g.}, grass $\rightarrow$ snow), object appearance changes (\textit{e.g.}, color), and object type/material substitutions (\textit{e.g.}, sheep $\rightarrow$ giraffe) while keeping the overall scene visually plausible. Captions under each image indicate the text prompt used to produce the corresponding edited sample, highlighting that the resulting manipulations are often photorealistic and view consistent, making detection based solely on 2D artifacts challenging.

\subsection{Real 3D Scenes}
Building a real-world benchmark for our task requires processing a large number of scenes, each with multi-view imagery, accurate camera poses, and a stable 3D scene representation for evaluation. Collecting and reconstructing such data from scratch would be prohibitively time-consuming and would introduce additional variability due to the reconstruction pipeline itself (\textit{e.g.}, pose estimation failures, incomplete coverage, or inconsistent scene quality). For this reason, we started from UCO3D~\cite{liu2025uncommon}, an existing large-scale dataset that already provided these prerequisites in a consistent format.

UCO3D contains a wide variety of objects, spanning more than $1000$ categories, captured in diverse indoor and outdoor environments, and additionally provides pretrained 3D Gaussian Splatting reconstructions. Leveraging these ready-to-use reconstructions allows scaling the dataset construction to many scenes while keeping reconstruction quality consistent across categories, so that our evaluation primarily reflects the behavior of the editing method rather than artifacts of the reconstruction process.

To avoid biasing the benchmark toward categories with many instances, we balance the subset across categories. Specifically, for each category we select
\[
n = \min_{c} \ \texttt{num\_objects}(c),
\]
\textit{i.e.}, the minimum number of available instances across categories, and sample $n$ objects per category.

\subsection{Fake 3D Scenes Generation} \label{sec:generation}
As mentioned, we employed two different methods to edit real scenes.

The first one is GaussCtrl~\cite{wu2024gaussctrl}, which edits rendered images from 3D Gaussian Splatting and recreates the edited scene through depth-conditioned editing based on ControlNet~\cite{zhang2023adding} for geometry consistency and attention-based latent code alignment for improving consistency during editing. 

The second one is Instruct-GS2GS~\cite{igs2gs} which uses InstructPix2Pix~\cite{brooks2023instructpix2pix} to iteratively edit the input images while optimizing the underlying scene, resulting in an optimized 3D scene edited accordingly to the instruction.

To construct our dataset, edited captions were automatically generated using an LLM, specifically the Meta-Llama-3-8B-Instruct~\cite{llama3modelcard}. We chose this model because it is open-source and can be deployed locally, allowing full control over generation parameters, reproducibility of the editing process, and large-scale caption generation without reliance on external APIs~\cite{catalini2025llms}. For each original caption, we create one edited caption by the following procedure: 
(i) we randomly sample one prompt template from the three defined below (the major differences between them are highlighted in italic);
(ii) we append a fixed suffix to the input caption with the goal to enforce the output format as a single caption only.
(iii) The resulting full prompt, obtained by concatenating the prompt, the caption, and the suffix, is then fed to the LLM, and its output is used as the edited caption.
For clarity and replicability, we report the prompts and the suffix used.

\begin{quote}
    \noindent \textbf{Prompt 1}: Modify the following sentence by changing \textit{the material or the type of the main object}, but do not change \textit{the color, the background, or the shape}.
\end{quote}

\begin{quote}
    \noindent \textbf{Prompt 2}: Modify the following sentence by changing \textit{the background or surface on which the main object stands}, but do not change \textit{the color, the shape, or any attribute of the main object}.
\end{quote}

\begin{quote}
    \noindent \textbf{Prompt 3}: Modify the following sentence by changing \textit{the color of the main object}, but do not change \textit{the shape or any other attribute of the main object}.
\end{quote}

\begin{quote}
    \noindent \textbf{Suffix}: The output should be a single caption without any additional explanation or text.
\end{quote}

Three edited captions were generated for each scene, corresponding to the three different prompts. For each scene, one caption was randomly selected and assigned, ensuring a balanced distribution of edit types across the dataset.

\subsection{Benchmark} \label{sec:benchmark}

We adopt different dataset splitting strategies depending on the evaluation setting.
In all cases, splits are performed at the edit level rather than enforcing a strict scene-level separation, allowing different edited versions of the same underlying 3D scene to appear across partitions.

\begin{enumerate}
    \item \textbf{Mixed split.} In the mixed setting we perform an $80-20$\% train--test split, resulting in approximately $33$k scenes for training and $8$k scenes for testing. This setting evaluates overall fake detection performance in a classic machine learning setup in which all training data (in this case, the generators) are known and available at training time.

    \item \textbf{Cross-editing split.} To evaluate generalization across editing methods, we define two additional splits where the model is trained on fake scenes generated with one editing approach and tested on the other (and vice versa). This setup ensures that the manipulation method used at test time is unseen during training, reflecting a more realistic scenario in which newly developed editing techniques are deployed after the detection model has already been trained. In this manner, we investigate the performance of a detector against unseen generators.
    
\end{enumerate}

\section{Experimental Results}

\subsection{Baselines}
Since, to the best of our knowledge, no method specifically addresses fake detection for 3D Gaussian scenes, we adopt state-of-the-art 2D deepfake detectors as baselines and evaluate them on renderings of the corresponding 3DGS scenes. In addition to our method, we compare against established detectors designed to recognize synthetic images produced by modern generators.

In our experiments, all baselines are fine-tuned on our training split (starting from the authors' released weights) to ensure a fair comparison under the same data distribution. Specifically, we include the models of Wang \textit{et al.}~\cite{wang2020cnn}, which build on a ResNet-50 backbone~\cite{NEURIPS2024_ebc62a3a} and investigate training strategies based on different image transformations to improve robustness and cross-generator generalization. Furthermore, we evaluate the reconstruction-based approach of Wang \textit{et al.}~\cite{wang2023dire}, which detects synthetic content by comparing an input image to its reconstruction and analyzing the resulting residual, and we also include CoDE (Contrastive Deepfake Embeddings)~\cite{baraldi2024contrastive}, which learns a deepfake oriented embedding space via contrastive learning while enforcing global/local agreement between full-image and cropped-view representations to capture both global cues and fine-grained artifacts. Finally, we include CLIP-based baselines following Ojha \textit{et al.}~\cite{ojha2023towards}: we load pretrained CLIP encoders~\cite{radford2021clip} and train a linear classifier on top of the CLIP image embedding. In this setting, we report results for two CLIP variants (ViT-B/16 and ViT-L/14), finetuned on our training split under the same protocol as the other baselines. Additionally, we evaluate self-supervised visual backbones based on DINOv2~\cite{oquab2024dinov2}, by fine-tuning the corresponding pretrained representations with an identical linear classification head.

For all experiments involving 2D detectors, we render images from each 3D scene and fine-tune/evaluate the baselines on these renderings using the same split protocol described in Section~\ref{sec:benchmark}.

\subsection{Proposed method}
We propose a 3D fake detector, here referred to as \textit{Fake3DD}, that operates directly on the Gaussian splatting representation. As backbone, we select the PointTransformerV3 model~\cite{wu2024ptv3}, a transformer-based architecture for point clouds that treats a point set as an unordered collection of 3D samples and learns features by local self-attention: for each point, the model aggregates information from nearby points in space. This design is well-suited to our setting because a 3DGS scene can be naturally viewed as a set of primitives distributed in 3D, where the authenticity cues may depend on spatial context rather than on individual Gaussians in isolation. 

Inspired by the work of Wu et al.~\cite{wu2024ptv3} and Li et al.~\cite{li2025scenesplat}, we applied some modifications to the original PointTransformerV3 model to enable the processing of Gaussian splats as input primitives instead of standard point clouds. These changes mainly concern the input feature representation, which is extended to incorporate the attributes associated with each Gaussian. Beyond these adaptations, we benefit from the inherent flexibility of the adopted architecture, which naturally supports batching scenes with varying numbers of Gaussian splats. This property allows us to avoid padding or resampling strategies and enables efficient training and inference across heterogeneous scenes. 

We propose to represent each Gaussian through the 3D coordinates of its mean point, and instead of the traditional color feature used for point clouds, we provide opacity, scale, quaternions, spherical harmonics $s_h$ with the zeroth-order term $s_0$ representing the view-independent color component. Both the encoder and decoder stages of PointTransformerV3 are used to extract contextualized Gaussian-level features. Given that each 3D scene is represented by a variable number of Gaussian splats, these Gaussian-level features are aggregated into a single scene-level representation using a global mean pooling operation. Specifically, features belonging to the same scene are averaged based on their batch indices, producing a fixed-dimensional embedding per scene. These embeddings are then passed to a classification head for the final binary prediction (fake or real classes).

\begin{table}[!t]
\centering
\caption{Experimental results on Fake3DGS dataset under different train/test split protocols. We report overall accuracy together with class-wise accuracy on fake and real samples. For each backbone, we evaluate (i) cross-edit generalization by training on one editing method (GaussCtrl or Instruct-GS2GS) and testing on the other, and (ii) the mixed setting where training and testing use the combined data.}
\label{tab:finetune_results}
\setlength{\tabcolsep}{3pt}
\scriptsize
\begin{tabular}{lccccc ccc}
\toprule
 & \multicolumn{2}{c}{\textbf{Train}} & \multicolumn{2}{c}{\textbf{Test}} &  & \multicolumn{3}{c}{\textbf{Accuracy} (\%)} \\
\cmidrule(lr){2-3}\cmidrule(lr){4-5}\cmidrule(lr){7-9}
Backbone & GaussCtrl & Instruct-GS2GS & GaussCtrl & Instruct-GS2GS &  & Overall & Fake & Real \\
\midrule
CLIP ViT-B & \cmark &        &        & \cmark &  & 80.7 & 70.6 & 95.9 \\
CLIP ViT-B &        & \cmark & \cmark &        &  & 74.3 & 41.5 & 98.5 \\
CLIP ViT-B & \cmark & \cmark & \cmark & \cmark &  & 84.0 & 84.2 & 93.8 \\
CLIP ViT-L & \cmark &        &        & \cmark &  & 81.5 & 67.1 & 96.6 \\
CLIP ViT-L &        & \cmark & \cmark &        &  & 75.4 & 44.5 & 98.2 \\
CLIP ViT-L & \cmark & \cmark & \cmark & \cmark &  & 84.0 & 83.7 & 94.3 \\
DINOv2-B   & \cmark &        &        & \cmark &  & 76.8 & 72.7 & 89.2 \\
DINOv2-B   &        & \cmark & \cmark &        &  & 73.2 & 45.9 & 93.3 \\
DINOv2-B   & \cmark & \cmark & \cmark & \cmark &  & 87.8 & 89.7 & 85.8 \\
\midrule
CoDE~\cite{baraldi2024contrastive}       & \cmark &        &        & \cmark &  & 80.2 & 60.5 & 92.4 \\
CoDE~\cite{baraldi2024contrastive}       &        & \cmark & \cmark &        &  & 79.3 & 54.1 & 95.8 \\
CoDE~\cite{baraldi2024contrastive}       & \cmark & \cmark & \cmark & \cmark &  & 92.2 & 91.4 & 92.9 \\
DM~\cite{corvi2023detection}         & \cmark &        &        & \cmark &  & 83.8 & 74.8 & 95.8 \\
DM~\cite{corvi2023detection}         &        & \cmark & \cmark &        &  & 82.4 & 80.3 & 96.9 \\
DM~\cite{corvi2023detection}         & \cmark & \cmark & \cmark & \cmark &  & 90.4 & 89.3 & 97.2 \\
UFD~\cite{ojha2023towards}        & \cmark &        &        & \cmark &  & 79.5 & 69.4 & 91.9 \\
UFD~\cite{ojha2023towards}        &        & \cmark & \cmark &        &  & 78.2 & 59.5 & 95.4 \\
UFD~\cite{ojha2023towards}        & \cmark & \cmark & \cmark & \cmark &  & 89.9 & 90.6 & 89.2 \\
\midrule
\textbf{Fake3DD} (Ours)       &        & \cmark & \cmark &        &  & 98.2 & 97.8 & 98.6 \\
\textbf{Fake3DD} (Ours)      & \cmark &        &        & \cmark &  & 98.6 & 99.1 & 98.3 \\
\textbf{Fake3DD} (Ours)      & \cmark & \cmark & \cmark & \cmark &  & \textbf{98.9} & \textbf{98.1} & \textbf{99.5} \\
\bottomrule
\end{tabular}
\end{table}

\subsection{Results}

Table~\ref{tab:finetune_results} summarizes the performance of 2D baselines and our 3D Gaussian-based detector under mixed and cross-edit protocols. In the mixed setting (both editors in train and test), most 2D detectors achieve reasonably high overall accuracy, with the strongest baseline reaching 92.2\% (CoDE). However, when evaluated under cross-edit generalization, all 2D methods exhibit a pronounced drop. Importantly, this degradation is largely driven by failures on the \emph{Fake} class: \textit{e.g.}, CLIP and DINO variants can maintain high \emph{Real} accuracy (often above 93--98\%) while their \emph{Fake} accuracy collapses (down to 41.5-45.9\% in several cases).

Notably, the strongest cross-edit baseline is DM trained on GaussCtrl and tested on Instruct-GS2GS ($G{\rightarrow}I$), achieving 83.8\% overall accuracy. However, the class-wise results indicate that this performance is still driven by a much higher accuracy on real samples (95.8\%) than on fake samples (74.8\%). This gap suggests that, under an unseen editor, the detector remains conservative and tends to misclassify a substantial portion of edited scenes as real.

In contrast, our method reaches 98.7\% on the same $G{\rightarrow}I$ protocol and remains well balanced across classes (99.1\% on fake and 98.4\% on real). Compared to the best cross-edit baseline, this corresponds to a +14.9 percentage points gain in overall accuracy, largely explained by a +24.3 pp improvement on the fake class (74.8\% $\rightarrow$ 99.1\%), while also improving real accuracy (+2.6 pp). These results support the hypothesis that 2D detectors tend to exploit editor-specific cues that do not transfer across manipulation pipelines, whereas operating directly on the 3DGS representation provides more editor-agnostic evidence for authenticity.

\begin{figure}[h]
    \centering
    \includegraphics[page=1, width=0.9\linewidth]{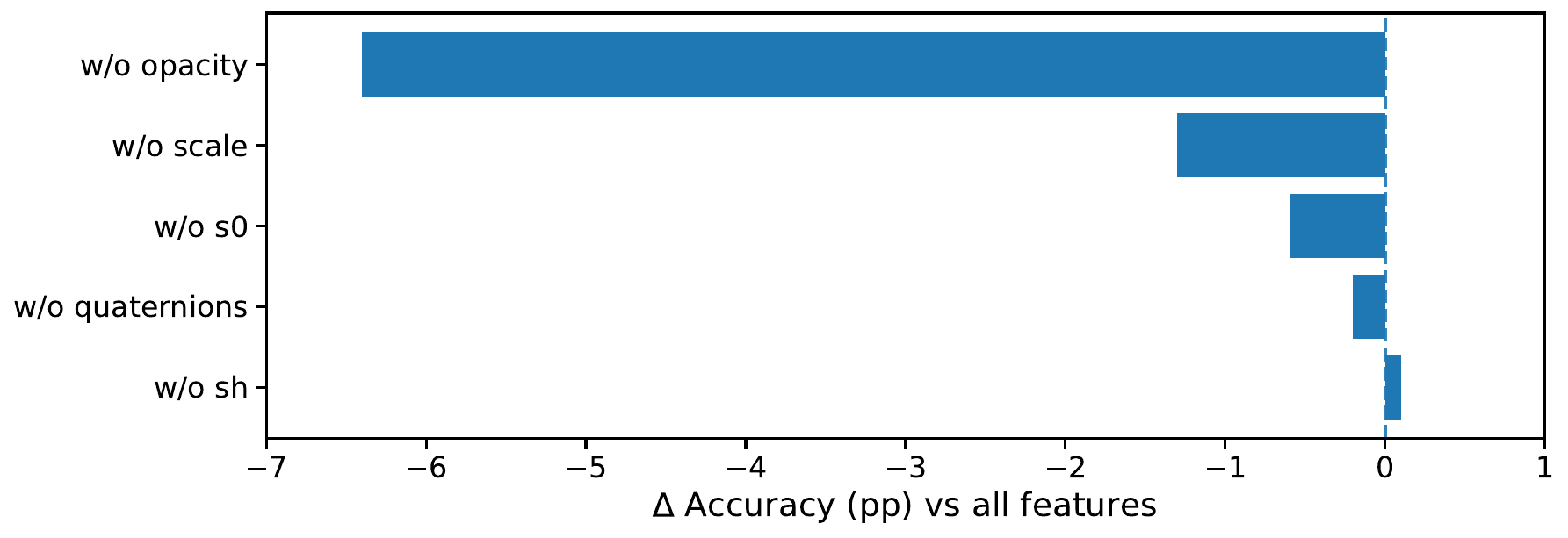}
    \caption{Ablation over Gaussian feature groups. We report the change in accuracy (in percentage points) obtained by removing each feature group from the input, relative to the full-feature model.}
    \label{fig:ablations}
\end{figure}

\subsection{Ablation Study}

As shown in Figure~\ref{fig:ablations}, we conducted an ablation study to assess the contribution of each Gaussian attribute to the detection performance. 

We remove one feature group at a time from the input and report the resulting change in accuracy with respect to the full model. Removing opacity caused the largest drop in accuracy (92.5\%), indicating that transparency information is critical. Scale and the zeroth-order spherical harmonics coefficient  $s_0$ provide a moderate contribution to detection performance. Scale captures geometric properties related to the spatial extent and distribution of Gaussian primitives, which can be affected by scene editing. The  $s_0$ term encodes view-independent color information; however, its impact is limited due to redundancy with other attributes and the scene-level mean pooling operation. Quaternion-based orientation features exhibit minimal influence on performance, suggesting that Gaussian orientation remains largely consistent between real and fake scenes and is therefore less informative for this task and carry limited discriminative information. Higher-order SH coefficients had minimal impact suggesting that view-dependent appearance cues are largely irrelevant for this binary classification task. Despite removing $s_h$ seems to improve the overall accuracy of 0.1, this was not the case in the cross setting, resulting in a lower accuracy w.r.t using $s_h$.  We therefore retain $s_h$ in all main experiments to favor robustness and generalization across editing methods.

\begin{figure}[h]
    \centering
    \includegraphics[page=2, width=1\linewidth]{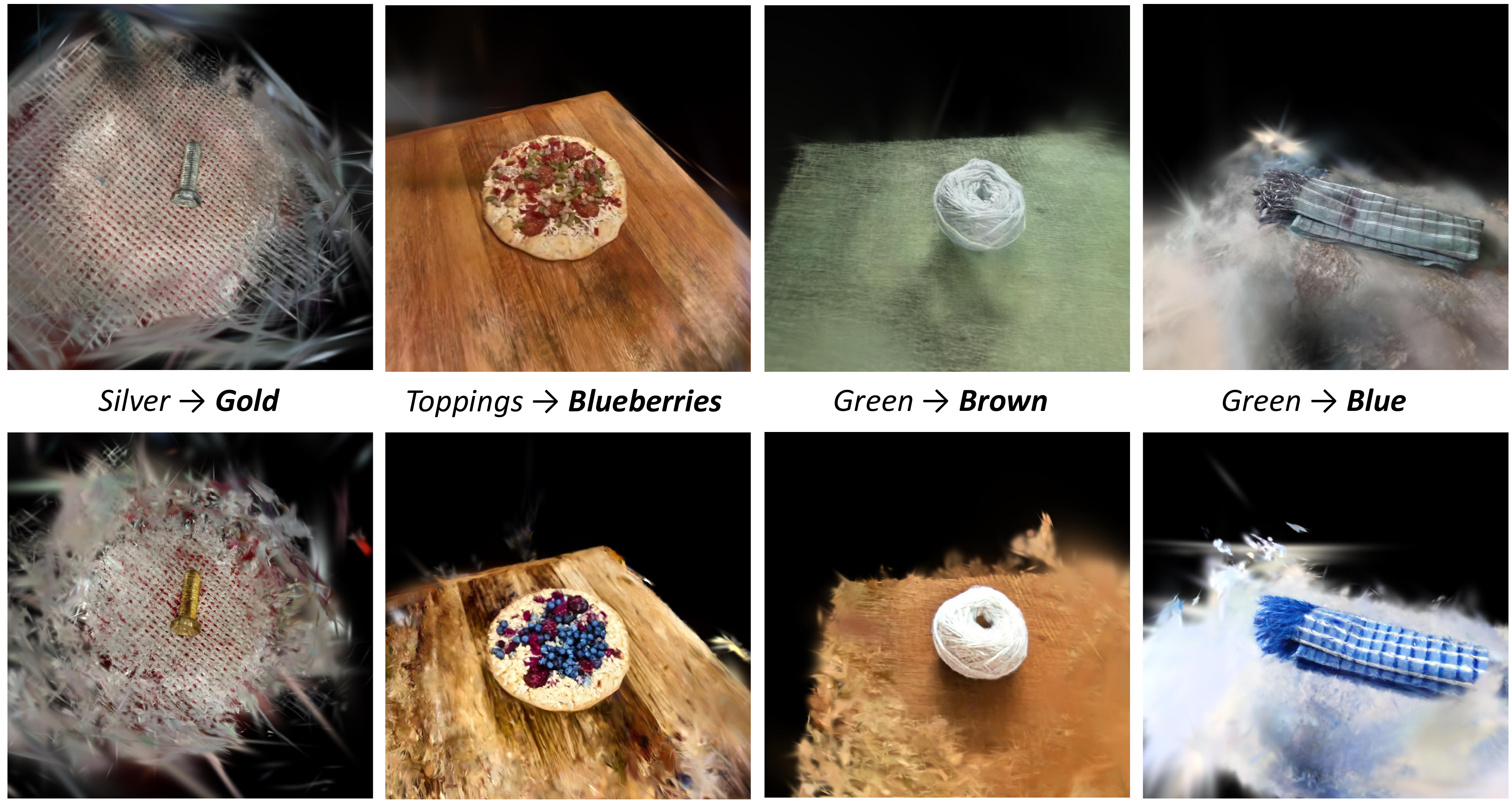}
    \caption{Some sample renderings of the data in our dataset. Below each image, there is the correspondent prompt used to generate edited samples.}
    \label{fig:renderings3D}
\end{figure}
\section{Conclusions and Future Work}
In this work, we introduced the important -- but still underexplored -- concept of 3D fake detection, which currently represent a security threat.
To further investigate this novel task, we presented \method, a large-scale benchmark composed of over $40$k real and edited 3D scenes generated using different editing methods. We also proposed a novel 3D fake detection framework based on Gaussian splatting representations. By leveraging Gaussian-level attributes and a modified PointTransformerV3 architecture, our approach effectively captures scene-level inconsistencies introduced by editing operations. 

Experimental results demonstrate that while 2D-based fake detection methods perform well when trained and tested on similar editing distributions, they struggle to generalize across unseen editing methods. In contrast, our 3D approach consistently outperforms 2D baselines, particularly in cross-edit evaluation settings, highlighting its robustness to editor-specific artifacts and its ability to generalize to novel manipulations. An extensive ablation study further shows that opacity and geometric attributes play a crucial role in detection, whereas view-dependent appearance cues contribute marginally to performance.

Future work will focus on evaluating the proposed framework using a broader range of 3D editing methods to assess its generalization capabilities further. Additionally, we plan to expand the dataset by explicitly annotating the spatial location and extent of edits within each scene, enabling more fine-grained tasks such as edit localization and region-level fake detection. We believe these directions will help advance the study of reliable and robust fake detection in 3D content generation pipelines.

{\small
\bibliographystyle{splncs04}
\bibliography{main}}

\end{document}